# Extreme AutoML: Analysis of Classification, Regression, and NLP Performance

Edward Ratner
Verseon International Corp.
Fremont, California
eratner@verseon.com

Brandon Warner
Verseon International Corp.
Fremont, California
bwarner@verseon.com

Amaury Lendasse
Dept. of Engineering Management and Systems Engineering
Missouri University of Science and Technology
Rolla, Missouri
alendasse@mst.edu

Elliot Farmer
Verseon International Corp.
Fremont, California
efarmer@verseon.com

Christopher Douglas
Verseon International Corp.
Fremont, California
cdouglas@verseon.com

*Abstract*— Utilizing machine learning techniques has always required choosing hyperparameters. This is true whether one uses a classical technique such as a KNN or very modern neural networks such as Deep Learning. Though in many applications, hyperparameters are chosen by hand, automated methods have become increasingly more common. These automated methods have become collectively known as automated machine learning, or AutoML. Several automated selection algorithms have shown similar or improved performance over state-of-the-art methods. This breakthrough has led to the development of cloud-based services like Google AutoML, which is based on Deep Learning and is widely considered to be the industry leader in AutoML services. Extreme Learning Machines (ELMs) use a fundamentally different type of neural architecture, producing better results at a significantly discounted computational cost. We benchmark the Extreme AutoML technology against Google's AutoML using several popular classification data sets from the University of California at Irvine's (UCI) repository, and several other data sets observing significant advantages for Extreme AutoML in accuracy, Jaccard Indices, the variance of Jaccard Indices across classes (i.e. class variance) and training times. (*Abstract*)

## I. INTRODUCTION

The selection of model structure and the corresponding hyperparameters is a challenge faced by modern practitioners of machine learning on a regular basis. This applies to methods that have been around for many years such as KNN's where the number of neighbors to be used and the choice of the distance metric need to be selected, as well modern Deep Learning networks where the number of hidden layers, the number of neurons per layer, the choice of activation functions and regularization have to all be decided prior to training the model (Ivakhnenko et al., 1967). In this paper, we focus on the application of extreme learning machines (ELMs), where the choices of the number of neurons and regularization needs to be made among others (Miche et al., 2009; Khan et al., 2020; Liu and Wang, 2010; Van Heeswijk et al., 2011; Grigorievskiy et al., 2014). Many practitioners select hyperparameters based on their prior experience. Many employ a trial-and-error method using a training and validation set. More recently some practitioners have employed numerical optimization techniques in order to optimize the hyperparameters. In that case, the parameters of the optimization method itself become new hyperparameters that are chosen manually. Each of these techniques require human involvement.

There are also automated ways to select optimal hyperparameters. These methods are often referred to as AutoML (Yao et al., 2018; He et al., 2021; Waring et al., 2020). Several automated selection algorithms have shown similar or improved performance over state-of-the-art methods (Zoph et al., 2016; Pham et al., 2018; Snoek et al., 2012). Google AutoML is considered by most to be the leader of this field. While Google AutoML employs Deep Learning in their models, Extreme AutoML (Warner et al., 2024) uses an ensemble of Extreme Learning Machines (ELMs) to operate without the need to set hyperparameters. Several ELM ensemble techniques have been proposed for varying application domains (Song et al., 2018; Yu et al., 2014; Han et al., 2015; Lan et al., 2009). Some earlier 58 comparisons between Google Auto ML and Extreme AutoML were reported by our team (Warner et al., 2023; Khan et al., 2023). We extend that analysis in this work. We chose four popular classification datasets from the UCI repository, measuring the methods' performance based on accuracy, Jaccard Indices, class variance, and training times. We also benchmark the performance of Extreme AutoML on a regression problem and an NLP classification problem. Section 2 provides an overview of Extreme Learning Machines. In section 3, we outline our novel approach to AutoML technology. In section 4, we provide an overview of the structured datasets and benchmarking methodology. In section 5, we present the results for the structured datasets. In section 6, we discuss the application of Extreme AutoML to the problem of movie revenue prediction. Section 7 covers the application of Extreme AutoML to an NLP data set. In section 8, we summarize our findings.

*A. Extreme Learning Machines*

Extreme Learning Machines (ELMs) are a form of generalized feedforward neural networks that employ a single layer of randomly generated hidden neurons. These types of networks can solve classification, regression, and clustering problems. ELMs are a member of the class of neural architecture called Randomized Neural Networks (RNNs). These networks contrast many other common architectures in that they randomly generate hidden nodes. Thus, weights from the first layer may be independent from the training data. Previous research has shown that ELMs are able to provide a unified learning platform with widespread feature mappings in both regression and classification tasks (Huang et al., 2004). Furthermore, ELMs are more easily optimizable when compared to classical algorithms like Support Vector

Machines (SVMs) due to the single-layer architecture (Lendasse et al., 2013).

ELMs are fundamentally different from networks that rely on conventional back-propagation due to the lack of dependence between input and output weights. Thus, ELMs have a non-iterative, linear ordinary least squares solution for the output weights. Huang et al. (2004) illustrated ELM's universal approximation capability, suggesting ELMs can universally approximate any continuous target functions in any compact subset X in Euclidean space. ELMs have illustrated learning speeds thousands of times faster than other feedforward neural network algorithms that rely on backpropagation while also obtaining greater generalization performance (Huang et al., 2004; Warner et al., 2023). ELMs may use any activation function that is an infinitely differentiable function (i.e., monotonic) and is bounded from -1 to 1 (e.g., hyperbolic 89 tangent). The output of an ELM may be denoted as

$$y_k = \sum_{j=1}^{m} \beta_{j,k}\, g\left(\sum_{i=1}^{n} w_{i,j} x_i + b_j\right)$$

where $x$ represents the input to the network and $y$ as the k'th output. $n$, $m$, and $k$ represent the number of neurons in the input, hidden, and output layers, respectively (Huang et al., 2004). $w$, and $\beta$, represent the corresponding weights for the input and output neurons. $b$ represents the bias values used 96 by the neurons in the hidden layer. Lastly, $g(.)$ denotes the activation function employed for each neuron. Eq. (1) may then be written as

$$\mathbf{H}\beta = y$$

where H is the hidden layer output matrix. The output neurons' weights, β1···m,1···k, are then computed with the generalized Moore–Penrose pseudoinverse matrix as

$$\beta = \mathrm{H}^+ y,$$

where H+ is the generalized Moore–Penrose pseudoinverse matrix of H, and y is the resultant output of the system. There are various modeling decisions to consider when using ELMs. First, one must choose an activation function that is monotonic and is bounded from -1 to 1 (e.g., hyperbolic tangent. Theoretically, one could scan different activation functions but this is generally held constant. The hyperparameters of ELMs include the number of neurons, the value of the bias term, weights, and regularization term (alpha). While the weights are random, one must choose a method for selecting a diverse combination of random weights. Additionally, if one is using an ensemble of ELMs, one must decide on the number of ELMs in the ensemble. Just as in other neural architectures, one may select the optimal hyperparameters using iterative, hybrid, or grid search approaches. In sum, conventional neural architectures that utilize backpropagation suffer from significant bottlenecks caused by slow gradient-based learning algorithms and iterative hyperparameter tuning. ELMs, on the other hand, provide state-of-the-art predictive performance using a fraction of the compute resources required by Deep Learning architectures due to the randomized single-hidden layer structure. As illustrated in the experiments below, this results in the ELM's ability to provide similar or superior performance in a small fraction of the time taken by Deep Learning.

## II. MATERIALS AND METHODS

### A. Overview of Novel Approach

Unlike Deep Learning architectures, a single ELM does not need a computationally intensive, iterative process to determine its neuron weights. Our recent paper introduces our novel ensemble learning methodology (Warner et al., 2024). In this approach, the neuron weighs are determined by solving a single linear system. An ensemble of ELMs can then be created to achieve more robust performance. Though many ensembling techniques have been explored, the choice of hyperparameter selection for each ELM in the ensemble remains an open question. Extreme AutoML has approached machine learning classification, regression, and time-series prediction in a fundamentally different way, where a proprietary algorithm is used to choose each ELMs hyperparameters (Warner et al., 2023). This allows our Extreme AutoML technology to produce fast, accurate models on both small data and large sets. This novel architecture enables the users of our technology to leverage their dynamic data in real-time.

Classical and Deep Learning algorithms also require extensive hyperparameter optimization. Extreme AutoML's revolutionary architecture automatically produces highly accurate models without the need to iteratively optimize hyperparameters, allowing users to instead focus their efforts on leveraging these state-of-the-art models to their advantage. Another important consideration to consider when choosing the best machine learning technology for classification problems is the variability of performance across classes. While a model with an overall relatively high accuracy may seem adequate at first, poor performance in one or more of the classes will result in a significant decrease in utility. The metric that best measures performance across classes is the Jaccard Index (Fletcher and Islam, 2018):

$$J(A, B) = \frac{|A \cap B|}{|A \cup B|}$$

where J is the Jaccard Distance between sets A and B, where the Jaccard Index is computed for each class. The Jaccard index is a monotonically increasing function of the F1-score, and therefore, both criteria would lead to the same rankings (Fletcher and Islam, 2018). Many models produce higher Jaccard indices for over-represented class, while the underrepresented classes have significantly lower Jaccard indices. This results in high Jaccard Index variability – a highly undesirable result in most applications. Models that produce higher Jaccard Indices for underrepresented classes and thus much lower Jaccard index variability illustrate greater generalization and utility. In the following sections, we report the results for classification. Since the results reported earlier (Warner et al., 2023), the Extreme AutoML platform has undergone coding optimization resulting in further improvements in efficiency.

### B. Structured Datasets and Methodology

The University of California at Irvine's machine learning repository contains thousands of open-source datasets used for classification, regression, time series, and clustering experiments. To best benchmark Extreme AutoML's technology against state-of-the-art industry leaders, performance comparisons were made on classification models side by side with Google's AutoML service. Classification is a machine learning task by which an algorithm attempts to correctly label a given validation sample based on the sample's characteristics. These classification models are trained in a supervised manner, meaning that the components of the model are learned based on examples whose classes are known. Metrics used to measure a classification model's performance include accuracy and Jaccard Indices for each class in the dataset. In some cases, both a training set and a test set were provided. In those cases, the training set was used to train the models and the test for validation. This was done to stay consistent with the goals of the originators of the benchmark data sets. In other cases, the data was not split in the benchmark. In those cases, k-fold cross-validation was used as the most accurate way of assessing machine learning technology.

The first experiment utilized the Human Activity Recognition Using Smartphones dataset (Anguita et al, 2013), which was separated into predefined training testing splits (80% training, 20% testing) and consisted of attributes and 10,299 observations. The second dataset, Parkinson's Disease Classification (Sakar et al., 2019), was not split beforehand. Therefore, 5-fold cross-validation was used (five splits of 80% training, and 20% testing). The original Parkinson's dataset is 184 variables by 756 observations, but since Google AutoML requires a minimum of 1000 observations to train a model, the dataset was doubled, and 0.1% noise was added to the "repeated" samples. The third experiment used the QSAR Oral Toxicity (Ballabio et al., 2019) dataset, also using 5-fold cross-validation. However, Google AutoML failed to train on two of the five folds, so only the successfully predicted folds were included in the comparison. The fourth dataset, CNAE-9 (Ciarelli and Oliveira, 2012), failed to train using Google AutoML, so the dataset was doubled, and 0.1% of noise was 190 added to the training sets. Only one of the five folds was successfully trained using Google AutoML, so only that fold was included in the comparison. In running Extreme Auto ML, the algorithm has a fast and accurate mode. The accurate method generates a search grid with 10 times the number of points as the fast method in the hyperparameter space.

### III. RESULTS

#### A. Classification

*a) Human Activity Recognition Using Smartphones*

Table 1 shows the breakdown of the Human Activity Recognition (Anguita et al., 2013) dataset's attributes and the performance for each learning method, each using one training set per each of the 8239 samples. The data in the Human Activity Recognition set were collected using 30 volunteers equipped with smartphones (Samsung Galaxy S II). While the smartphone was mounted on the participants' waist, the volunteers were recorded while performing common activities of daily living (ADL) including walking, walking upstairs, walking downstairs, sitting, standing, and laying down. The embedded sensors in the smartphone (accelerometer and gyroscope) then recorded 3-axial linear acceleration and 3-axial angular velocity acceleration measurements, which maintained at a constant rate of 50Hz. The sensor signals were then pre-processed using noise filters, sampling in fixe width sliding windows of 2.56 sec and 50% overlap. The acceleration signals, which contain gravitational and body motion variables, were separated using a Butterworth low-pass filter. Since gravitational forces only contain low frequency components, a filter cutoff of 0.3 Hz was used. Finally, each window measurement resulted in a vector of features, which were obtained by calculating variables from time and frequency domains. Figure 2 shows the various Jaccard Indices for the six classes in the Human Activity Recognition dataset. For each class, Extreme AutoML outperformed Google AutoML in terms of Jaccard Indices, showing greater generalization capabilities.

**Table 1.** Human Activity Recognition results

| Method | Training time (seconds) | Accuracy |
|---|---|---|
| **Extreme AutoML** | 21 | 0.9626 |
| **Google AutoML** | 14700 | 0.9460 |

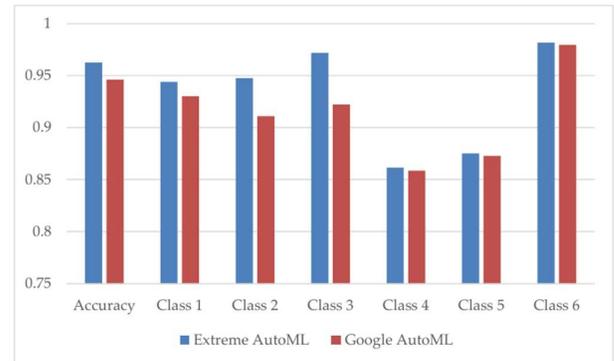

**Figure 1.** Classification Performance – Human Activity Recognition dataset.

*b) Parkinson's Disease Classification*

In the second experiment using the Parkinson's dataset (Sakar et al., 2019), 1202 training samples were used for the 752 attributes. These data were collected using patients (23 men and 41 women) with Parkinson's disease, ranging in ages from 33 to 87. The data collection process began by setting the microphone to 44.1 KHz, followed by a physician's examination consisting of pronouncing the sustained phonation of the vowel /a/ in three repetitions. Preprocessing steps were conducted using various speech signal processing algorithms including Time Frequency Features, Mel Frequency Cepstral Coefficients (MFCCs), Wavelet Transform based Features, Vocal Fold Features and TWQT features. In our experiment, the Extreme AutoML accurate method outperformed Google AutoML by 1.45% accuracy whilst training almost 300 times faster. Figure 3 shows the breakdown of the accuracy and Jaccard Indices for the two classes in the dataset. For each of the classes and methods, Extreme AutoML outperformed Google AutoML in terms of Jaccard Indices, further illustrating greater generalization capabilities.

**Table 2.** Parkinson's classification performance

| Method | Training time (seconds) | Accuracy |
|---|---|---|
| **Extreme AutoML (fast)** | 21.5 | 0.9088 |
| **Extreme AutoML (accurate)** | 215 | 0.9233 |
| **Google AutoML** | 59820 | 0.9046 |

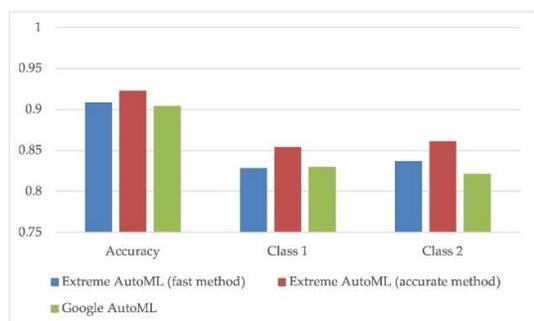

**Figure 2.** Parkinson's prediction results

*c) QSAR Oral Toxicity*

The third experiment used data from the QSAR Oral Toxicity dataset (Ballabio et al., 2019). This dataset was developed at the Milano Chemometrics and QSAR Research Group in Bicocca, Italy, in collaboration with the U.S. Environmental Protection Agency. The researchers used a set of chemicals provided by the ICCVAM Acute Toxicity Workgroup (U.S. Department of Health and Human Services) to develop in-silico models for the purpose of predicting oral systemic toxicity for filling regulatory needs. In this experiment, AutoML accurate method performed 2.17% better than Google AutoML whilst training 24 times faster. Table 5 shows the breakdown of the Jaccard Indices for each of the classes in the QSAR Oral Toxicity model. For each of the classes and methods, Extreme AutoML outperformed Google AutoML in terms of Jaccard Indices. Most notably, in this unbalanced dataset, the Jaccard Index for the underrepresented class was more than 2x higher for the Extreme AutoML model than the Google AutoML model.

**Table 3.** QSAR Oral Toxicity results

| Method | Training time (seconds) | Accuracy |
|---|---|---|
| **Extreme AutoML (fast)** | 157 | 0.937 |
| **Extreme AutoML (accurate)** | 1570 | 0.9394 |
| **Google AutoML** | 37260 | 0.9177 |

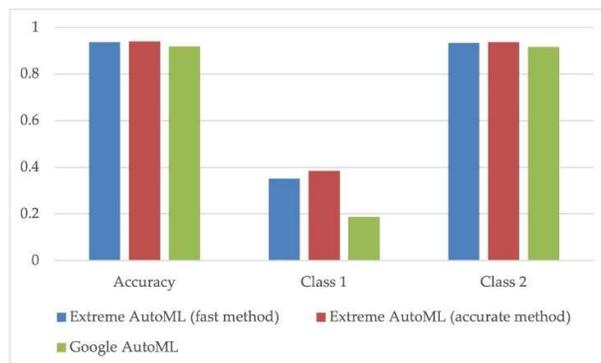

**Figure 3.** QSAR Oral Toxicity results

*d) CNAE-9 Business Classiciation*

The CNAE-9 dataset (Ciarelli and Oliveira, 2012) contains vectorized representations of 108 documents of free-text business descriptions of Brazilian companies organized into a subset of 9 categories catalogued in the National Classification of Economic Activities database. Preprocessing steps conducted by the researchers included removing prepositions, transforming words into their canonical form, and vectorizing each document where the weight of each word is its frequency in the document. This dataset is highly sparse, with over 99% of the matrix being filled with zeros. Using this data set, our Extreme AutoML approach performed 1.4% better than Google AutoML, whilst training 1760 times faster. Figure 5 shows the breakdown of the Jaccard Indices for 269 each of the classes in the CNAE9 Business Classification model. Our AutoML's minimum Jaccard Index was higher than Google AutoML by 0.1, a 20% improvement.

**Table 4.** CNAE-9 classification results

| Method | Training time (seconds) | Accuracy |
|---|---|---|
| **Extreme AutoML** | 7.8 | 0.7721 |
| **Google AutoML** | 13740 | 0.7581 |

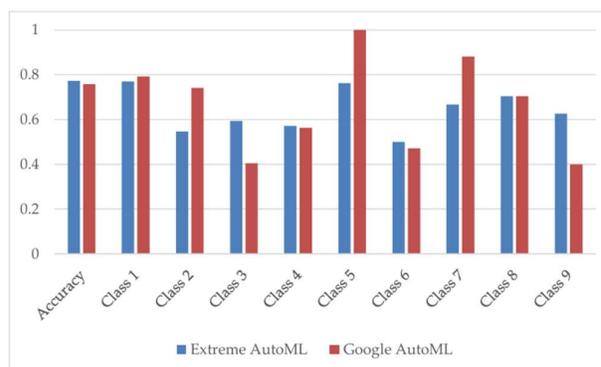

**Figure 4.** Classification performance on CNAE-9

*e) Classification Results Summary*

In this section, we share the overall classification performance of the two methodologies by comparing relative error rates, training times, and variance (both relative and absolute) of Jaccard Indices across classes for all datasets. We begin by presenting the relative error rates in Figure 5 below.

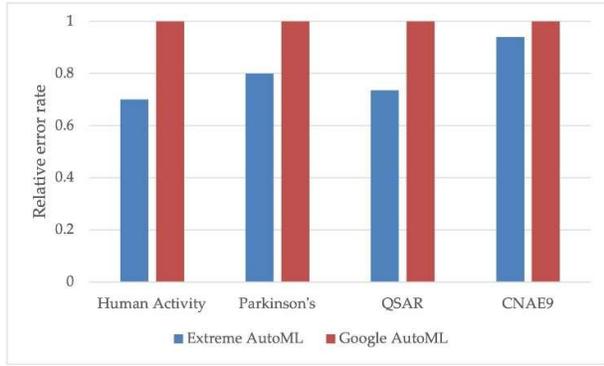

**Figure 5.** Classification results summary

Figure 6 illustrates the difference in the absolute variance of the Jaccard Indices for all methodologies and datasets.

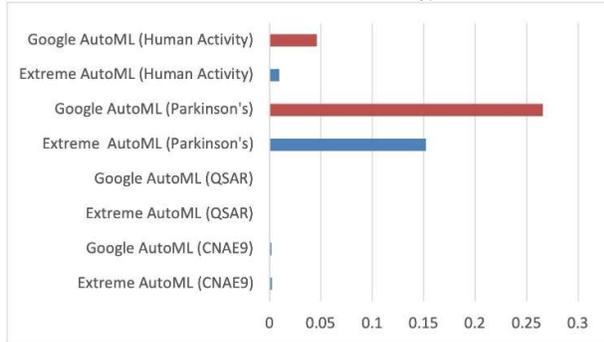

**Figure 6.** Absolute variance of Jaccard Indices

Extreme AutoML produced a lower variance of Jaccard Indices across classes for each of the datasets when compared to Google AutoML in each case when the class variance was significant. This indicates that the performance of the Extreme AutoML classifier is significantly more uniform across the dataset in addition to being more accurate. Examining the results in more detail, we see that Extreme AutoML Jaccard index is significantly higher than that of Google AutoML for the strongly underrepresented classes. This has significant implications in many important applications. This further illustrates Extreme AutoML's greater generalization capabilities. Figure 7 visualizes the vast disparity in training times between our novel approach and Google AutoML. Note, the training times are shown on a logarithmic scale because our approach would not be visible without it.

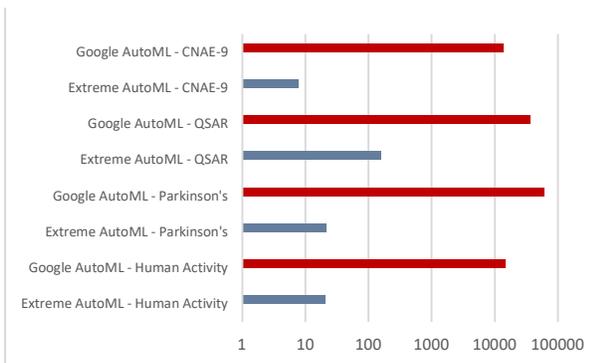

**Figure 7.** Summary of training times (logarithmic scale)

## IV. REGRESSION RESULTS

### B. Regression Results

#### a) Movie Revenue Prediction

The movie industry generates billions of dollars of revenue annually. The industry has, however, historically struggled with predicting which movies will end up being successful and which will end up flopping. Given the large investment needed to produce modern movies, it is highly desirable to understand the likely outcome of any given movie release. To that end, several researchers have attempted to predict movie success as defined by various success metrics. Several studies have attempted to predict movie popularity, normally 307 partitioned into a few categories (Vijarania et al., 2022; Lakshmi et al., 2020; Iqbal et al., 2021; Sahu 308 et al., 2023; Mbunge et al., 2022). From a business planning point of view, predicting the actual revenue generated by a movie is significantly more desirable. This, up to this point, has proven quite difficult. Though a number of publications report results for movie revenue prediction (Mbunge et al., 2022; Zhang et al., 2009; Zhou et al., 2019; Zhou et al., 2018), these do not actually attempt to predict the numerical revenue. Rather, the attempt to classify the movie revenue into a handful of classes, which is a much more limited type of information. Even with that constraint, the best results have relatively modest, with the best current outcomes for American movies standing at around 55% accuracy for classification into six classes (Zhou et al., 2018). Here, we report the application of Extreme Auto ML technology to predict the revenue of American movies. This was motivated by both commercial considerations, as well as our desire to demonstrate Extreme AutoML on a full regression problem. The data was extracted from the very popular IMDB website. The data sets contain information such as the movie budget, the movie genre, the release date, the full list of the cast by name, and the full list of the crew by name. The data set also contains the generated revenue and the IMDB popularity, which were both measured after the movie release. For this study, we focused on revenue prediction and discarded the popularity data. The non-numerical inputs were one-hot encoded and individuals who appeared in the list only a few times were discarded. Our final data set consisted of 3049 movies for which both the production budget and the final revenue were available.

**Table 5.** Movie revenue prediction results

| Method | Pearson (R) |
|---|---|
| **Extreme AutoML** | 0.82 |
| **Google AutoML** | 0.70 |

We performed 5-fold cross-validation to characterize the performance of the Extreme AutoML models. We compute the final correlation on the result of the whole data set predictions. For baselines, we compare our results to the results of research conducted by Anderson et. al (2019). They obtained their best correlation results using the XGBoost regressor. As can be seen from Table 9, Extreme AutoML significantly outperforms the XGBoost regressor. Further, the results obtained by our Extreme AutoML model are qualitatively good enough to generate further commercial interest.

## V. NATURAL LANGUAGE PROCESSING (NLP) RESULTS

### A. SMS Spam Classification

To demonstrate Extreme AutoML's suitability for natural language processing problems, we selected a dataset of 5572 SMS texts from the UCI ML Repository. This corpus is a collection of SMS 342 spam messages, which were manually extracted from the Grumbletext website (a UK forum where users submit text messages, they receive which they believe might be spam). Another subset of 3,375 SMS randomly chosen ham messages from the NUS SMS Corpus (NSC) were added to the corpus. These messages were collected manually by the Department of Computer Science at the National University of Singapore. These messages were mostly produced by Singaporean students attending the university. The messages were collected from volunteers who were knowingly aware that their contributions were to be made public. Additionally, a list of 450 SMS ham messages were collected from Caroline Tag's PhD Thesis. Finally, the corpus contained the SMS Spam Corpus v.0.1 Big, which contains 1,002 SMS ham messages and 322 spam messages. In this experiment, two Extreme AutoML classification models were trained using industry-standard feature extraction techniques – the first using BERT (Devlin et al., 2019) and the second using RoBERTa (Liu et al., 2019). These results were then compared to results obtained by using the OpenAI platform which uses GPT-3 (Brown et al., 2020), the state-of-the-art language model (at the time of publication) which powers ChatGPT. For reference, we also included classification benchmarks using a convolutional neural network trained with Word2Vec (Mikolov et al., 2013), a previous generation feature extraction technique, as well as a multinomial Naïve Bayes model trained using TF-IDF (Sammut et al., 2010). For these less sophisticated methods, SMOTE oversampling (Chawla et al., 2002) was utilized to artificially bring 360 the target "spam" class from approximately 13% up to 50%.

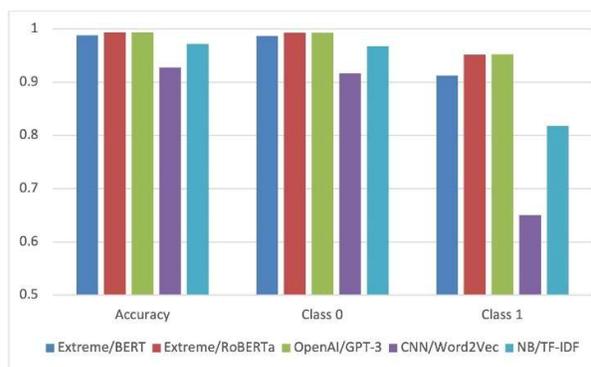

**Figure 7.** NLP results on SMS text classification data set

Extreme AutoML, while only having access to the publicly available BERT and RoBERTa feature extraction technologies, was able to achieve comparable results with OpenAI, despite the latter having the benefit of using next-generation GPT-3 feature extraction. Especially when paired with RoBERTa. Extreme AutoML's Jaccard index for the heavily underrepresented Class 1 demonstrates its remarkable prowess in NLP applications.

## VI. CONLUSIONS AND FUTURE WORK

In this paper, we compare Extreme AutoML to state-of-the-art approaches for various machine learning contexts. We compare the classification performance on several classification datasets from the UCI repository using Google AutoML and Extreme AutoML. We compare performance with a regression problem of movie revenue prediction between Extreme AutoML and XGBoost. We also analyzed the performance of Extreme AutoML on an NLP dataset, comparing the results to OpenAI, as well as several more traditional methodologies. We also provide an overview of Extreme AutoML's novel approach, differentiating it from the near-ubiquitous Deep Learning paradigm. We show significantly better results on a number of key metrics of interest for the Extreme AutoML approach for classification problems, regression problems, and NLP applications. In future work, we plan to do additional benchmarking across an even wider range of machine-learning applications.

## VII. AUTHOR CONTRIBUTIONS

E. R., E. F., B. W., and C. D. are all employed by Verseon International Corp. A. M. is a consultant to Verseon International Corp.

## VIII. FUNDING

This research received no external funding.

## IX. ACKNOWLEDGEMENTS

The content of this manuscript has been presented, in part, at the IEEE Extreme Learning Machine 2021 Conference, (Warner, et al., 2023).

## X. DATA AVAILABILITY STATEMENT

The regression dataset analyzed for this study can be found in the Film Success GitHub repository [https://github.com/ryan-anderson-560ds/explorations/blob/master/film_success/tmdb_5000_movies.csv]. The rest of the data is available from the UCI Repository.

## XI. REFERENCES

Akusok, A., Björk, K.M., Estévez, V. and Boman, A., 2021, December. Randomized Model Structure Selection Approach for Extreme Learning Machine Applied to Acid Sulfate Soil Detection. In International Conference on Extreme Learning Machine (pp. 32-40). Cham: Springer International Publishing.

Akusok, A., Miche, Y., Hegedus, J., Nian, R. and Lendasse, A., 2014. A two-stage methodology using K-NN and false-positive minimizing ELM for nominal data classification. Cognitive Computation, 6, pp.432-445.

Anguita, D., Ghio, A., Oneto, L., Parra, X. and Reyes-Ortiz, J.L., 2012. Human activity recognition on smartphones using a multiclass hardware-friendly support vector machine. In Ambient Assisted This is a provisional file, not the final typeset article Living and Home Care: 4th International Workshop, IWAAL 2012, Vitoria-Gasteiz, Spain, December 3-5, 2012. Proceedings 4 (pp. 216-223). Springer Berlin Heidelberg.

Ballabio, D., Grisoni, F., Consonni, V. and Todeschini, R., 2019. Integrated QSAR models to predict acute oral systemic toxicity. Molecular informatics, 38(8-9), p.1800124.

Brown, T., Mann, B., Ryder, N., Subbiah, M., Kaplan, J.D., Dhariwal, P., Neelakantan, A., Shyam, P., Sastry, G., Askell, A. and Agarwal, S., 2020. Language models are few-shot learners. Advances in neural information processing systems, 33, pp.1877-1901.

Cambria, E., Huang, G.B., Kasun, L.L.C., Zhou, H., Vong, C.M., Lin, J., Yin, J., Cai, Z., Liu, Q., Li, K. and Leung, V.C., 2013. Extreme learning machines [trends & controversies]. IEEE intelligent systems, 28(6), pp.30-59.

Carolus Khan, K., Ratner, E., Douglas, C. and Lendasse, A., 2021, December. A Novel Methodology for Object Detection in Highly Cluttered Images. In International Conference on Extreme Learning Machine (pp. 10-23). Cham: Springer International Publishing.

Chawla, N.V., Bowyer, K.W., Hall, L.O. and Kegelmeyer, W.P., 2002. SMOTE: synthetic minority over-sampling technique. Journal of artificial intelligence research, 16, pp.321-357.

Ciarelli, P. and Oliveira, E. (2012). CNAE-9. UCI Machine Learning Repository. https://doi.org/10.24432/C51G7P.

Devlin, J., Chang, M.W., Lee, K. and Toutanova, K., 2018. Bert: Pre-training of deep bidirectional transformers for language understanding. arXiv preprint arXiv:1810.04805.

Dua, D.; Graff, C. UCI Machine Learning Repository, http://archive.ics.uci.edu/ml. Irvine, CA: University of California, School of Information and Computer Science, 2019.

Fletcher, S. and Islam, M.Z., 2018. Comparing sets of patterns with the Jaccard index. Australasian Journal of Information Systems, 22.

Grigorievskiy, A., Miche, Y., Ventelä, A.M., Séverin, E. and Lendasse, A., 2014. Long-term time series prediction using OP-ELM. Neural Networks, 51, pp.50-56.

Han, B., He, B., Nian, R., Ma, M., Zhang, S., Li, M. and Lendasse, A., 2015. LARSEN-ELM: Selective ensemble of extreme learning machines using LARS for blended data. Neurocomputing, 149, pp.285-294.

He, X., Zhao, K., & Chu, X. (2021). AutoML: A survey of the state-of-the-art. Knowledge-Based Systems, 212, 106622.

https://github.com/ryan-andersonds/explorations/tree/master/film_success UCI Machine Learning Repository: SMS Spam Collection Data Set.
https://archive.ics.uci.edu/ml/datasets/SMS+Spam+Collection (accessed 2023-02-14).

Huang, G. B., Zhu, Q. Y., & Siew, C. K. (2004, July). Extreme learning machine: a new learning scheme of feedforward neural networks. In 2004 IEEE international joint conference on neural networks (IEEE Cat. No. 04CH37541) (Vol. 2, pp. 985-990). IEEE.

Iqbal, N., Ahmad, R., Jamil, F. and Kim, D.H., 2021. Hybrid features prediction model of movie quality using Multi-machine learning techniques for effective business resource planning. Journal of Intelligent & Fuzzy Systems, 40(5), pp.9361-9382.

Ivakhnenko, A.G. and Lapa, V.G., 1967. Cybernetics and forecasting techniques.

Jin, H., Chollet, F., Song, Q. and Hu, X., 2023. Autokeras: An automl library for deep learning. Journal of Machine Learning Research, 24(6), pp.1-6

Khan, K., Ratner, E., Ludwig, R. and Lendasse, A., 2020, July. Feature bagging and extreme learning machines: machine learning with severe memory constraints. In 2020 International Joint Conference on Neural Networks (IJCNN) (pp. 1-7). IEEE.

Komer, B., Bergstra, J., & Eliasmith, C. (2014, July). Hyperopt-sklearn: automatic hyperparameter configuration for scikit-learn. In ICML workshop on AutoML (Vol. 9, p. 50). Austin, TX: Citeseer.

Lakshmi, M., Shastry, K.A., Sandilya, A. and Shekhar, R., 2020, October. A comparative analysis of Machine Learning approaches for Movie Success Prediction. In 2020 Fourth International Conference on I-SMAC (IoT in Social, Mobile, Analytics and Cloud)(I-SMAC) (pp. 684-689). IEEE.

Lan, Y., Soh, Y.C. and Huang, G.B., 2009. Ensemble of online sequential extreme learning machine. Neurocomputing, 72(13-15), pp. 3391-3395.

Lauren, P., Qu, G., Yang, J., Watta, P., Huang, G.B. and Lendasse, A., 2018. Generating word embeddings from an extreme learning machine for sentiment analysis and sequence labeling tasks. Cognitive Computation, 10, pp.625-638.

Lendasse, A., He, Q., Miche, Y. and Huang, G.B., 2014. Advances in extreme learning machines (ELM2012). Neurocomputing, (128), pp.1-3.

Liang, J., Meyerson, E., Hodjat, B., Fink, D., Mutch, K. and Miikkulainen, R., 2019, July. Evolutionary neural automl for deep learning. In Proceedings of the Genetic and Evolutionary Computation Conference (pp. 401-409).

Lipton, Z.C., Elkan, C. and Naryanaswamy, B., 2014. Optimal thresholding of classifiers to maximize F1 measure. In Machine Learning and Knowledge Discovery in Databases: European Conference, ECML PKDD 2014, Nancy, France, September 15-19, 2014. Proceedings, Part II 474 14 (pp. 225-239). Springer Berlin Heidelberg.

Liu, N. and Wang, H., 2010. Ensemble based extreme learning machine. IEEE Signal Processing Letters, 17(8), pp.754-757.

Liu, Y., Ott, M., Goyal, N., Du, J., Joshi, M., Chen, D., Levy, O., Lewis, M., Zettlemoyer, L. and Stoyanov, V., 2019. Roberta: A robustly optimized bert pretraining approach. arXiv preprint arXiv:1907.11692.

Mbunge, E., Fashoto, S.G. and Bimha, H., 2022. Prediction of box-office success: A review of trends and machine learning computational models. International Journal of Business Intelligence and Data Mining, 20(2), pp.192-207.

Miche, Y., Sorjamaa, A., Bas, P., Simula, O., Jutten, C. and Lendasse, A., 2009. OP-ELM: optimally pruned extreme learning machine. IEEE transactions on neural networks, 21(1), pp.158-162.

Miche, Y., Van Heeswijk, M., Bas, P., Simula, O. and Lendasse, A., 2011. TROP-ELM: a double-regularized ELM using LARS and Tikhonov regularization. Neurocomputing, 74(16), pp.2413-2421.

Mikolov, T., Chen, K., Corrado, G. and Dean, J., 2013. Efficient estimation of word representations in vector space. arXiv preprint arXiv:1301.3781.

Pham, H., Guan, M., Zoph, B., Le, Q., & Dean, J. (2018, July). Efficient neural architecture search via parameters sharing. In International conference on machine learning (pp. 4095-4104). PMLR.

Radzi, S.F.M., Karim, M.K.A., Saripan, M.I., Rahman, M.A.A., Isa, I.N.C. and Ibahim, M.J., 2021. Hyperparameter tuning and pipeline optimization via grid search method and tree-based autoML in breast cancer prediction. Journal of personalized medicine, 11(10), p.978.

Sahu, S., Kumar, R., Long, H.V. and Shafi, P.M., 2023. Early-production stage prediction of movies success using K-fold hybrid deep ensemble learning model. Multimedia Tools and Applications, 82(3), pp.4031-4061.

Sakar, C.O., Serbes, G., Gunduz, A., Tunc, H.C., Nizam, H., Sakar, B.E., Tutuncu, M., Aydin, T., Isenkul, M.E. and Apaydin, H., 2019. A comparative analysis of speech signal processing algorithms for Parkinson's disease classification and the use of the tunable Q-factor wavelet transform. Applied Soft Computing, 74, pp.255-263.

Sammut, C. and Webb, G.I. eds., 2011. Encyclopedia of machine learning. Springer Science & Business Media.

Snoek, J., Larochelle, H. and Adams, R.P., 2012. Practical bayesian optimization of machine learning algorithms. Advances in neural information processing systems, 25.

Song, Y., Zhang, S., He, B., Sha, Q., Shen, Y., Yan, T., Nian, R. and Lendasse, A., 2018. Gaussian derivative models and ensemble extreme learning machine for texture image classification. Neurocomputing, 277, pp.53-64.

Stankovic, M., Bacanin, N., Zivkovic, M., Jovanovic, D., Antonijevic, M., Bukmira, M. and Strumberger, I., 2022, October. Feature Selection and Extreme Learning Machine Tuning by Hybrid Sand Cat Optimization Algorithm for Diabetes Classification. In International Conference on Modelling and Development of Intelligent Systems (pp. 188-203). Cham: Springer Nature 512 Switzerland.

Sui, X., He, S., Vilsen, S. B., Teodorescu, R., & Stroe, D.: Hyperparameter Optimization in Bagging-514 Based ELM Algorithm for Lithium-Ion Battery State of Health Estimation. In 2023 IEEE Applied Power Electronics Conference and Exposition (APEC), 2023, 1797-1801.

Thornton, C., Hutter, F., Hoos, H.H. and Leyton-Brown, K., 2013, August. Auto-WEKA: Combined selection and hyperparameter optimization of classification algorithms. In Proceedings of the 19th ACM SIGKDD international conference on Knowledge discovery and data mining (pp. 847-855).

Van Heeswijk, M., Miche, Y., Oja, E. and Lendasse, A., 2011. GPU-accelerated and parallelized ELM ensembles for large-scale regression. Neurocomputing, 74(16), pp.2430-2437.

Vijarania, M., Gambhir, A., Sehrawat, D. and Gupta, S., 2022. Prediction of movie success using sentimental analysis and data mining. In Applications of Computational Science in Artificial Intelligence (pp. 174-189). IGI Global.

Waring, J., Lindvall, C., & Umeton, R. (2020). Automated machine learning: Review of the state-of-the-art and opportunities for healthcare. Artificial intelligence in medicine, 104, 101822.

Warner, B., Ratner, E. and Lendasse, A., 2021, December. Edammo's Extreme AutoML Technology–Benchmarks and Analysis. In International Conference on Extreme Learning Machine (pp. 152-163). Cham: Springer International Publishing.

Warner, B., Ratner, E., Carlous-Khan, K., Douglas, C. and Lendasse, A., 2024. Ensemble Learning with Highly Variable Class-Based Performance. Machine Learning and Knowledge Extraction, 6(4), pp.2149-2160.

Wu, L., Perin, G. and Picek, S., 2022. I choose you: Automated hyperparameter tuning for deep learning-based side-channel analysis. IEEE Transactions on Emerging Topics in Computing.

Yao, Q., Wang, M., Chen, Y., Dai, W., Li, Y.F., Tu, W.W., Yang, Q. and Yu, Y., 2018. Taking human out of learning applications: A survey on automated machine learning. arXiv preprint arXiv:1810.13306.

Yu, Q., Miche, Y., Séverin, E. and Lendasse, A., 2014. Bankruptcy prediction using extreme learning machine and financial expertise. Neurocomputing, 128, pp.296-302.

Yu, Q., Van Heeswijk, M., Miche, Y., Nian, R., He, B., Séverin, E. and Lendasse, A., 2014. Ensemble delta test-extreme learning machine (DT-ELM) for regression. Neurocomputing, 129, pp.153-158.

Zhang, L., Luo, J. and Yang, S., 2009. Forecasting box office revenue of movies with BP neural network. Expert Systems with Applications, 36(3), pp.6580-6587.

Zhou, Y., Zhang, L. and Yi, Z., 2019. Predicting movie box-office revenues using deep neural networks. Neural Computing and Applications, 31, pp.1855-1865.

Zhou, Y. and Yen, G.G., 2018, July. Evolving deep neural networks for movie box-office revenues prediction. In 2018 IEEE Congress on Evolutionary Computation (CEC) (pp. 1-8). IEEE.

Zoph, B. and Le, Q.V., 2016. Neural architecture search with reinforcement learning. arXiv preprint arXiv:1611.01578.